\documentclass[a4paper, 10pt, conference]{ieeeconf}      % Use this line for a4
\usepackage{FG2023}
\usepackage{graphicx}
\usepackage{epsfig} % for postscript graphics files
\usepackage{mathptmx} % assumes new font selection scheme installed
\usepackage{times} % assumes new font selection scheme installed
\usepackage{amsmath} % assumes amsmath package installed
\usepackage{amssymb}  % assumes amsmath package installed

\FGfinalcopy % *** Uncomment this line for the final submission

\IEEEoverridecommandlockouts
\pubid{\makebox[\columnwidth]{979-8-3503-4544-5/23/\$31.00~\copyright~2023 IEEE \hfill}\hspace{\columnsep}\makebox[\columnwidth]{ }}

\def\FGPaperID{127} % *** Enter the FG2023 Paper ID here

\title{\LARGE \bf
Generalized Face Anti-Spoofing 
via Multi-Task Learning \\ 
and One-Side Meta Triplet Loss
}

\author{\parbox{16cm}{\centering
    {\large Chu-Chun Chuang$^1$, Chien-Yi Wang$^2$,and Shang-Hong Lai$^{1,2}$}\\
    {\normalsize
    $^1$ National Tsing Hua University, Taiwan \quad
    $^2$ Microsoft AI R\&D Center, Taiwan}}
}

\begin{document}

\ifFGfinal
\thispagestyle{empty}
\pagestyle{empty}
\else
\author{Anonymous FG2023 submission\\ Paper ID \FGPaperID \\}
\pagestyle{plain}
\fi
\maketitle

\pubidadjcol

%%%%%%%%%%%%%%%%%%%%%%%%%%%%%%%%%%%%%%%%%%%%%%%%%%%%%%%%%%%%%%%%%%%%%%%%%%%%%%%%
%%%%%%%%%%%%%%%%%%%%%%%%%%%%%%%%%%%%%%%%%%%%%%%%%%%%%%%%%%%%%%%%%%%%%%%%%%%%%%%%
\begin{abstract}

With the increasing variations of face presentation attacks, model generalization becomes an essential challenge for a practical face anti-spoofing system. This paper presents a generalized face anti-spoofing framework that consists of three tasks: depth estimation, face parsing, and live/spoof classification. With the pixel-wise supervision from the face parsing and depth estimation tasks, the regularized features can better distinguish spoof faces. While simulating domain shift with meta-learning techniques, the proposed one-side triplet loss can further improve the generalization capability by a large margin. Extensive experiments on four public datasets demonstrate that the proposed framework and training strategies are more effective than previous works for model generalization to unseen domains.

\end{abstract}

%%%%%%%%%%%%%%%%%%%%%%%%%%%%%%%%%%%%%%%%%%%%%%%%%%%%%%%%%%%%%%%%%%%%%%%%%%%%%%%%
\section{INTRODUCTION}

Owing to the progressive development of face recognition techniques, the information securities are integral to recognition systems. Thanks to face anti-spoofing, the probability of accessing the systems with presentation attacks have strongly reduced.

Over the past few years, face anti-spoofing methods can be roughly divided into appearance-based and temporal-based methods. However, the performance of these works significantly degrades while the distributions between testing data and training data exist significant discrepancies (sensors, environments, and spoofing mediums). Therefore, domain generalization becomes significant while dealing with the face anti-spoofing task. Some of the previous works~\cite{Jia_2020_CVPR_SSDG,Shao_2019_CVPR,Shao_2020_AAAI,D2AM-AAAI2021,ANRL-MM2021,SSAN-M-CVPR2022} have employed domain generalization techniques to face anti-spoofing. Adversarial training and meta-learning frameworks were used to find generalized features among different domains, and the performance showed verified improvements. The way that meta-learning achieves generalization is keeping simulating domain shift scenarios during the training process. The model learns the ability of transfer between domains. In this paper, we conduct meta-learning process to improve generalization and strengthen the optimization of inter-domain feature distributions.

\begin{figure}[thpb]
\begin{center}
   \includegraphics[width=\linewidth]{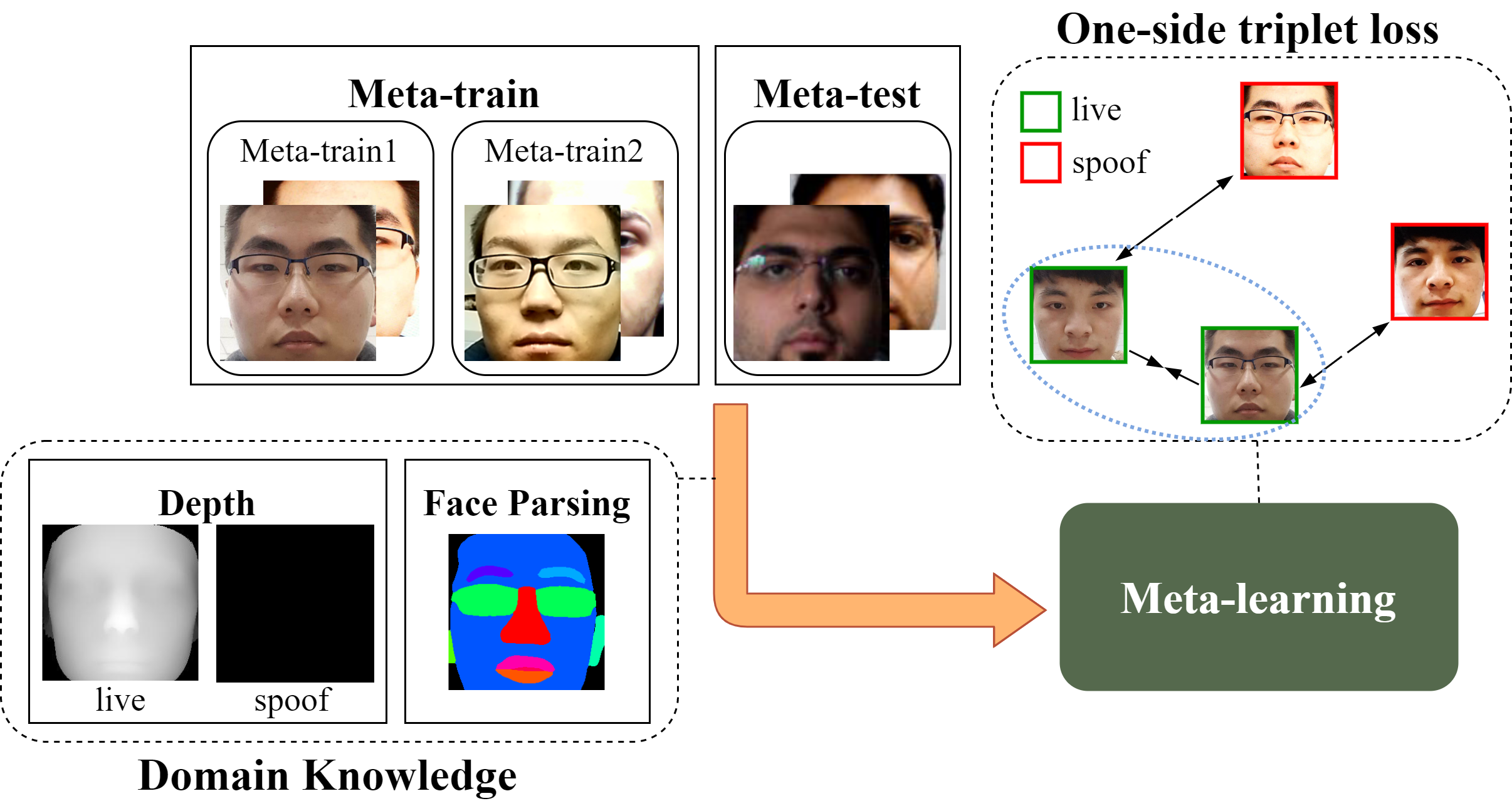}

   \end{center}
   \caption{Illustrated idea of the proposed face anti-spoofing system.}
\label{fig:intro}
\end{figure}

To find generalized features, employing multiple aspects of face information may help us achieve this goal. For example, the model would understand the face composition and depth conditions better with face parsing and depth information. Based on this idea, multi-task framework may be helpful to address this thought. In this way, the features our model learned would include various facial information. With understanding the key information of faces, we could find more generalized facial features. Moreover, spoof regions are not always the whole faces. There are many attack types of local regions such as eyes. Understanding face parsing would also help in this situation. Thus, here we employ face parsing to strengthen the face understanding capability and enrich there presentations that we learned. In the past, ~\cite{kalayeh2017,attributes,expression_synthesis} also employ face parsing to improve the understanding of faces. For the depth, it is one of the important facial information as well. Pixel-wise binary masks are employed in many works as supervision for training. ~\cite{Fang2021PartialAS} employs binary masks that determines the real region as pixel-wise supervision for partial attacks. Here we employ binary depth masks as pixel-wise supervision. Moreover, the depth maps of attacks like print and replay are flat. Thus, we encourage our model to understand faces much better and to regularize learning directions by leveraging depth facial features in this paper.

RFMetaFAS~\cite{Shao_2020_AAAI} conducts meta-learning process to improve generalized ability with domain shift simulations. It also employs depth information as the prior knowledge to regularize the learning direction. Inspired by RFMetaFAS~\cite{Shao_2020_AAAI} and our thoughts mentioned above, we propose a multi-task meta-learning framework that includes depth estimation, face parsing, and spoof classification tasks. For the face parsing task, we employ an U-net based face parsing module into the framework, which can serve as pixel-wise supervision and strengthen the 
understanding of different facial parts. However, meta-learning focuses on domain shift simulations which optimizes the inter-domain distributions. Hence, the distributions of intra-domain may not well constrained. At the same time, triplet loss can enforce similar data of different label to separate, and this is crucial due to the distribution of intra-domain is usually close. Moreover, as the spoof images are more dispersed due to the large variation of attack mediums and collection environments, the one-side triplet loss only samples triplet anchors from the features of live images. Hence, we propose to apply the one-side triplet loss in the meta-optimization procedure in the meta learner module. The one-side triplet loss encourages the aggregation of live features and is proven to be very effective for domain generalization in face anti-spoofing.    

%Pixel-wise supervision task like depth estimator~\cite{Liu_2018_CVPR} is proven to be effective for regularizing the learning direction of spoof classification.
%With the face parsing supervision and the attention-based skip connection, the anti-spoofing model can learn more generalized representations by leveraging semantic facial features. 

%Contributions
The main contributions of our work contain three aspects:

1. We introduce a multi-task meta-learning framework for learning more generalized features for face anti-spoofing. The framework and training techniques demonstrate superior performance over previous approaches in various domain generalization protocols of face anti-spoofing.

2. A U-net based face parsing module and the attention-based skip connection are utilized for encoding facial priors into the network and further promoting the feature generalization capability. 

3. During the meta-optimization procedure, we propose to use the one-side triplet loss, which encourages the aggregation of live features and discriminates the diverse spoof features better in unseen domains. 

%------------------------------------------------------------------------
\section{Related Works}

\subsection{Face Anti-spoofing}\label{s:antispoofing}
Due to the different aspects of exploiting face features, existing methods dealing with face anti-spoofing can be roughly divided into two groups: temporal-based methods and appearance-based methods.

\subsubsection{Temporal-based Methods}\label{s:temporal}
Temporal-based methods aim to exploit the temporal information from multiple frames of face videos to distinguish live and spoof faces. Recently, several temporally-based methods have been proposed. CNN-LSTM~\cite{Xu2015LearningTF} proposed to extract temporal features from multiple face frames with a CNN-LSTM architecture for anti-spoofing. Furthermore, ~\cite{Liu_2018_CVPR} exploited rPPG signals as useful information in face videos to distinguish spoof attacks from live faces.

\subsubsection{Appearance-based Methods}\label{s:appearane}
Comparatively, appearance-based methods aim to differentiate live and spoof faces from spatial information in images due to the different textures and features between live and spoof images. In recent years, CNN based methods ~\cite{yang2014learn} have been employed for face anti-spoofing. Two-stream CNN-based method~\cite{two-stream} combines patch-based CNN and depth-based CNN models for face anti-spoofing.

However, ~\cite{two-stream} learns the fused face representations via finding patch and depth features separately, whereas our work directly leverages face parsing and depth tasks to regularize the learning direction. The face representations extracted in our work contains both face parsing and depth information with one common feature extractor, and thus to obtain more generalized features.

Regardless of the tempotal-based or appearance-based methods mentioned above, the prior works are poor to generalize to unseen domains since data distributions are different from training data. Thus, several works including this paper exploit the concepts of domain generalization for face anti-spoofing. The next subsection we would introduce these works.

\subsection{Domain Generalization for Face Anti-Spoofing}\label{s:generalizarion}
Domain generalization(DG) aims to learn from several source domains and then test on the unseen target domain. Recently, domain generalization is an important challenge for face anti-spoofing as we have mentioned in introduction. Several learning methods such as adversarial learning and meta-learning are adopted to improve generalization. MADDG~\cite{Shao_2019_CVPR} uses a multi-adversarial discriminative deep domain generalization framework to learn generalized features from multiple domains. SSDG~\cite{Jia_2020_CVPR_SSDG} proposes single-side domain generalization for face anti-spoofing with single-side adversarial learning and an asymmetric triplet loss. RFMetaFAS~\cite{Shao_2020_AAAI} learns generalized features via meta-learning and employs depth information as prior knowledge. We deal with domain generalization of face anti-spoofing with also a meta-learning-based method. Different from others, our work employs multi-task framework to learn more complete and generalized facial features, and optimizes distributions with meta-learning and one-side triplet loss for inter-domain and intra-domain, separately.
\section{Proposed Method}
Our work aims to learn more robust features and improve the generalization capability of the face anti-spoofing model. Meta-learning is also effective for domain generalization due to domain shifting simulations. Here we propose a multi-task framework with one-side triplet loss during meta-optimization.

In the following section, we first introduce the proposed multi-task framework, followed by the components of the proposed framework: multi-task meta-learning, U-net based face parsing module, one-side triplet loss, and detailed objective functions.

\subsection{Overview}\label{s:overview}
\begin{figure*}[thpb]
\begin{center}
%\fbox{\rule{0pt}{2in} \rule{.9\linewidth}{0pt}}
    \includegraphics[width=0.9\textwidth]{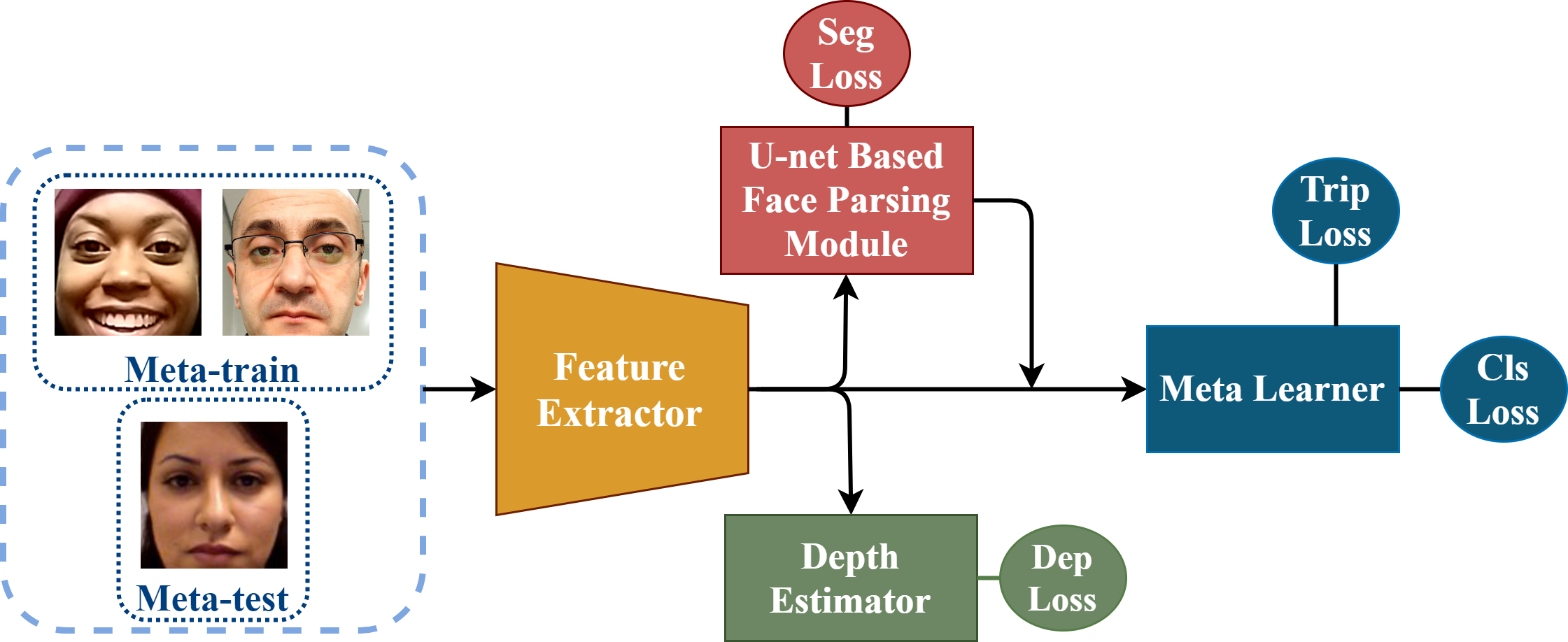}
\end{center}
   \caption{Our framework contains depth estimation, face parsing, and meta-learned spoof classification branches. Here we denote the feature extractor as $F$, U-net based face parsing module as $S$, depth estimator as $D$, and meta learner as $M$. Source images are divided into meta-train set and meta-test set in each iteration. The depth estimator $D$ is to predict depth maps, which serves as prior knowledge to regularize the model. U-net based face parsing module $S$ aims to improve the generalization by encoding the semantic facial priors with the attention-based skip-connection. In the end, features are fed into the meta learner $M$ to conduct meta-optimization, which involves the one-side triplet and classification losses.}
\label{fig:overview}
\end{figure*}

Fig. ~\ref{fig:overview} illustrates the overall framework. The proposed method is a multi-task meta-learning based approach, which contains a feature extractor, a meta learner for spoof classification, a depth estimator branch for depth estimation, and a U-net based face parsing module for semantic segmentation on faces. 

We assume each training data contains a face image \emph{x} associated with the corresponding label \emph{y}, face depth image \emph{d}, and face parsing image \emph{s}. For the network, we denote the feature extractor as $F$, U-net based face parsing module as $S$, depth estimator as $D$, and meta learner as $M$. In the network, the face image \emph{x} is fed into feature extractor $F$ to extract a feature vector \emph{f}. After the feature extraction, a depth estimator $D$ estimates the corresponding depth map \emph{d} from the feature vector \emph{f}. Simultaneously, the feature \emph{f} is fed into a U-net based face parsing module $S$, which helps the model learn better representations to obtain and strengthen the information of face parsing. Lastly, the aggregated feature vector, which is concatenated from feature \emph{f} and face parsing feature, is fed into the meta learner $M$ for live/spoof classification. Besides, we apply one-side triplet loss in meta learner to improve the data distribution in feature space. We explain the details in the coming sections.

\subsection{Multi-Task Meta Learning}\label{s:algo}
\begin{figure}[thpb]
\begin{center}
   \includegraphics[width=\linewidth]{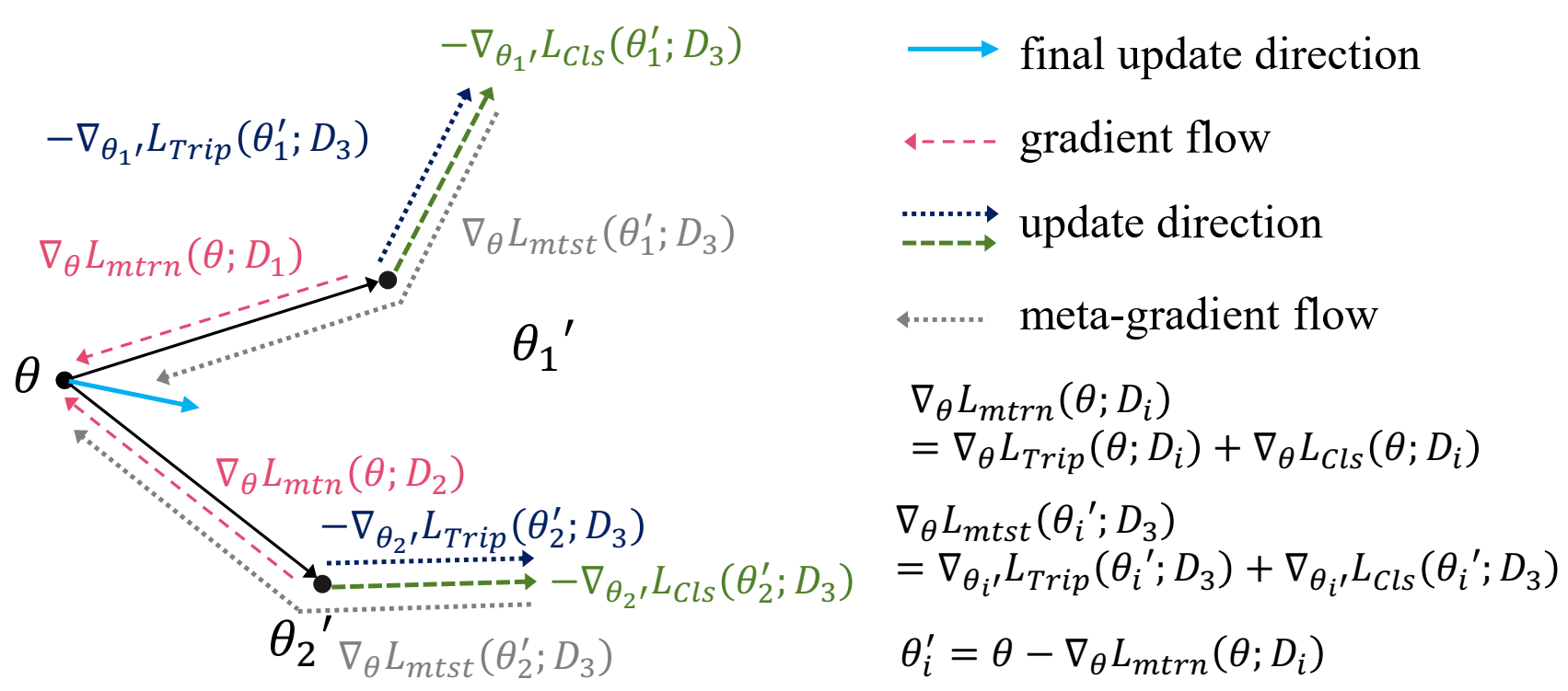}
\end{center}
   \caption{Illustration of the optimization process in meta learner. Given $D_1$, $D_2$ as meta-train domains, and $D_3$ as meta-test domain. $L_{mtrn}$, $L_{mtst}$, $L_{Cls}$ and $L_{Trip}$ denotes loss of meta-train, meta-test, spoof classification, and one-side triplet loss. $\theta_1'$ and $\theta_2'$ are found by $D_1$ and $D_2$, and $D_3$ is applied for finding learning directions by both $\theta_1'$ and $\theta_2'$. Every domain division contains a gradient and meta-gradient from meta-train and meta-test domains.}
\label{fig:gradient}
\end{figure}

We adopt the fine-grained learning strategy in RFMetaFAS~\cite{Shao_2020_AAAI} that we randomly choose one of the source domain as meta-test domain and the others are meta-train domains during each iteration. Here we assume N source domains in total, including N-1 meta-train domains and a meta-test domain. We assume $D_{mtrn}$ and $D_{mtst}$ as meta-train domains and meta-test domain, respectively. The meta-learning process contains meta-train, meta-test, and meta-optimization stages. The meta learner conducts meta-learning by exploiting a variety of domain shift scenarios in each iteration. Fig. ~\ref{fig:gradient} illustrates the detailed gradient flows of our meta-learning process.

\subsubsection{Meta-train}
For meta-train stage, data of each meta-train domain is sampled for batches. Here we exploit classification loss and one-side triplet loss to find learning directions. During the meta-train stage, the parameters of meta learner are first found by $\theta_{M_i'} = \theta_M-\alpha((\nabla_{\theta_M}L_{Cls(D_{mtrn}(i))}(\theta_F,\theta_M))+(\nabla_{\theta_M}L_{Trip(D_{mtrn}(i))}(\theta_F,\theta_M)))$ for each meta-train domain, where $\theta_F$ and $\theta_M$ denote the parameters of feature extractor and meta learner, $i\in{D_{mtrn}}$, $L_{Cls}$ and $L_{Trip}$ are the classification loss and one-side triplet loss, respectively. Besides, $L_{Dep}$ is adopted to regularize the feature extractor as well. For live faces, depth image is supposed to be a face-shape like depth maps, while depth images of spoof faces are all set to zero. Moreover, $L_{Seg}$ is adopted to learn face parsing information and find more generalized representations.

\subsubsection{Meta-test}
The generalization capability could be further enhanced with meta-learning by domain shifting simulations, and here meta-test domain acts as a role of the unseen domain. Thus, our model should perform well on the meta-test domain based on learning directions found by the meta-train domain. Parameters of the meta learner here are denoted as $\theta_{M_i'}$.

\subsubsection{Meta-Optimization}
In meta-optimization, we optimize our model, which includes $\theta_F$, $\theta_M$, $\theta_S$, and $\theta_D$, with the sum of meta-train and meta-test losses.

\subsection{U-net Based Face Parsing Module}\label{s:parsing}
\begin{figure}[t]
\begin{center}
   \includegraphics[width=\linewidth]{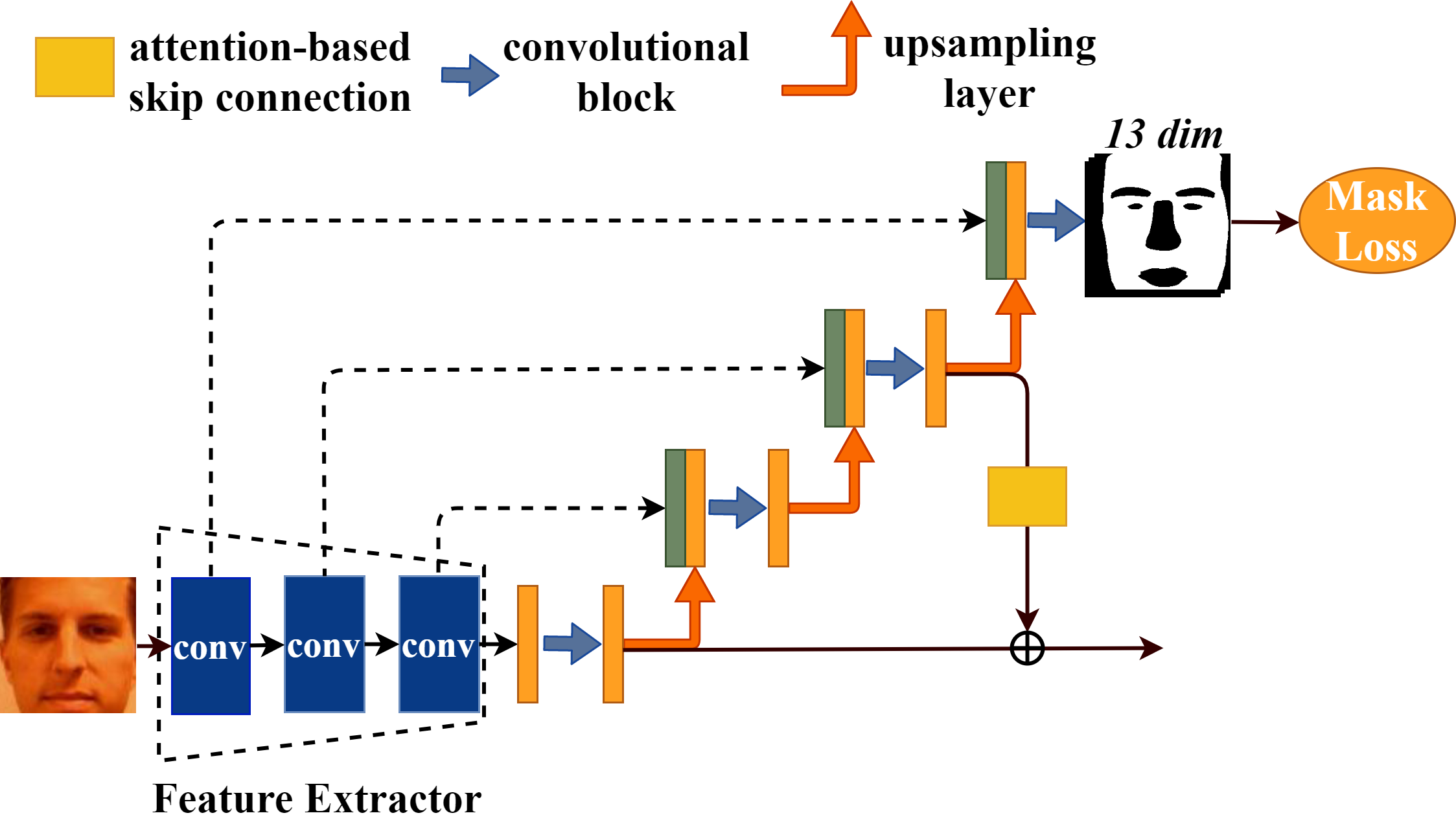}
\end{center}
   \caption{The architecture of U-net based face parsing module. Our face parsing module contains two parts: face parsing U-net and attention-based skip connection. The encoder of the U-net includes three convolutional blocks, which share weights with the feature extractor. Features after the encoder are fed into the decoder and upsampled three times, and there are skip connections to the corresponding convolutional block of the encoder. Apart from the output of 13 dimensions segmentation maps, there is another path called attention-based skip-connection, which contains a convolutional block and an ECA module to transfer important face parsing information to the main network.}
\label{fig:face parsing fig}
\end{figure}

U-net based face parsing module aims to learn semantic facial priors and discriminate the spoof images with the attention to different facial parts. Fig. ~\ref{fig:face parsing fig} shows the architecture of U-net face parsing module. Our face parsing module contains two parts: a face parsing U-net and an attention-based skip connection.

\subsubsection{Face Parsing U-net}\label{s:unet}
U-net is a commonly used architecture for dealing with semantic segmentation tasks. We employ a U-net based architecture that includes an encoder-decoder structure. Both encoder and decoder contain three convolutional blocks, and skip connections are added to aggregate segmented information between them. The face parsing U-net aims to predict face parsing images \emph{s}. The input of the face parsing module is the face image \emph{x}, which is the same as the feature extractor. Specifically, since we want face parsing information to regulate the feature extractor, we set the encoder and feature extractor with shared weights. After we obtain the feature \emph{f}, the decoder outputs the 13-dimension face parsing binary maps.

\subsubsection{Attention-Based Skip Connection}\label{s:branch}
We apply the attention-based skip connection to encode semantic facial priors into the main branch of the network. Here we take the last stage of the face parsing decoder as the input for the attention-based module. Due to the reason that different channels of the face parsing feature may exist specific connections between each other, here we apply a channel attention module ECA-Net \cite{eca-net} to learn inter-channel relations of the feature in the skip connection branch. Finally, the output feature is concatenated with feature \emph{f} for feeding into the meta learner.

\subsection{One-Side triplet loss}\label{s:learner}
Here we propose one-side triplet loss to combine the inter-domain advantage of meta learning and intra-domain one of triplet loss. First, triplet loss is suitable for face anti-spoofing due to aggregations of the same class and separations of different classes. In the meantime, meta learning aims to simulate domain shift and thus improves the generalization capability in unseen domains. However, meta learning mainly focuses on inter-domain distribution. Hence, we propose one-side triplet loss for both meta-train and meta-test domains. Besides, since meta learning emphasizes domain shifting directions for unseen domains, inter-domain triplet loss is redundant. 
Second, different from the normal triplet loss, one-side triplet only aggregates live domain data due to the reason below. Since there are countless kinds of attacks due to variations such as cameras, illuminations, etc., we believe that spoof data are more widely distributed than live ones. Moreover, since spoof domains could be kept dividing into smaller spoof subdomains, we believe distributions of different spoof datasets should be partially overlapped instead of completely independent to others. Thus, it is not necessary to separate different spoof domains. Separating live and spoof domains is a critical task. Therefore, different from SSDG~\cite{Jia_2020_CVPR_SSDG}, we propose one-side triplet loss, which only applies live domain for anchor and applies both live and spoof domains for positive and negative samples.
%\begin{figure}[h]
%    \centering
%    \includegraphics[width=0.9\columnwidth]{Triplet loss figure.png}
%    \caption{\label{fig:triplet loss fig}
%    Illustration of one-side triplet loss. One-side triplet loss combines the inter-domain advantage of meta learning and intra-domain one of triplet loss. It focuses on aggregating real domains and pulling off spoof domains but not aggregating spoof ones. Ultimately, we get the illustration of feature distributions like the right side.}
%\end{figure}

Due to the spoof domains are widely distributed, it is prone to pick extreme data while processing the one-side triplet loss. To avoid interference, the triplet loss mining strategy becomes pretty crucial. Hence, we apply two-stage mining triplet loss. A smaller margin is chosen at first and applies batch all mining strategy, optimizing all of the valid triplets. Then we increase the margin with the batch hard mining strategy, which optimizes only the hardest triplet. By doing this, we could get more stable results. Equation (\ref{trip_mtrn_eqn}) and (\ref{trip_tst_eqn}) show one-side triplet loss $L_{triplet}$ in meta-train and meta-test.

\subsection{Objective Function}\label{s:loss}
Our network is trained with four different objective functions, i.e., the classification loss, one-side triplet loss, segmentation loss, and depth loss. The four objective functions will be described in detail in this subsection. Here we assume $J$ denotes the domains belonging to $D_{mtrn}$, $J_i(i=1,2,...,N-1)$ and ${K}$ denotes the domain belonging to $D_{mtst}$.

\subsubsection{Classification Loss}\label{s:clsloss}
We adopt binary cross-entropy classification loss to both meta-train and meta-test domains for classification. Equation (\ref{cls_mtrn_eqn}) and (\ref{cls_tst_eqn}) give the classification losses during each meta-train and meta-test domain, respectively.
\begin{align}\label{cls_mtrn_eqn}
    L_{Cls( J_i)}( \theta_{F},\theta_{M})=&
    \sum_{(x,y) \sim{ J_i}}(y \log{M(F(x))} \nonumber\\
    + &(1-y) \log{(1-M(F(x)))})
\end{align}
\begin{align}\label{cls_tst_eqn}
    \sum \limits_{i=1}^{N-1} L_{Cls( {K})}( \theta_{F}, \theta_{M_i'}) =& \sum \limits_{i=1}^{N-1} \sum \limits_{(x,y) \sim {K}}(y \log{M_i'(F(x))} \nonumber\\
    + &(1-y) \log{(1-M_i'(F(x)))})
\end{align}
 where $\theta_{F}$, $\theta_{M}$, and $\theta_{M_i'}$ denote the parameters of feature extractor, meta learner, and meta learner after update from $i$-th domain in meta-train. $x$ and $y$ are the input and the corresponding label. 
\subsubsection{One-Side Triplet Loss}\label{s:triploss}
In order to combine with meta learning, we adopt one-side triplet loss into our method. The detailed equations of one-side triplet loss for each meta-train and meta-test domain are given in (\ref{trip_mtrn_eqn}) and (\ref{trip_tst_eqn}), respectively.
\begin{align}\label{trip_mtrn_eqn}
    L_{Trip(J_i)}(\theta_F,\theta_M) =& \sum_{(x_{i}^{a},x_{i}^p,x_{i}^n)}(\|{M(F(x_{i}^{a}))}-{M(F(x_{i}^{p}))}\|_{2}^{2} \nonumber\\
    - &\|M(F(x_{i}^{a}))-{M(F(x_{i}^{n}))}\|_{2}^{2}+\alpha)
\end{align}
\begin{align}\label{trip_tst_eqn}
    \sum&_{i=1}^{N-1}L_{Trip( {K})}( \theta_F, \theta_{M_i'}) =  \sum \limits_{i=1}^{N-1} \sum \limits_{(x_t^a,x_t^p,x_t^n)}(\|{M_i'(F(x_t^a))} \nonumber\\
    - &{M_i'(F(x_t^p))}\|_2^2-\|M_i'(F(x_t^a))-{M_i'(F(x_t^n))}\|_2^2+ \alpha)
\end{align}
where $x_i\sim{J}_i$ and $x_t\sim{K}$; $x^a$, $x^p$, $x^n$ denote anchor, positive, negative sample, respectively, and $\alpha$ is a margin.

\subsubsection{Segmentation Loss}\label{s:segloss}
Since we expect the U-net based face parsing module to predict the face parsing masks corresponding to the source face image, here we adopt the multi-class cross-entropy loss. The segmentation masks are 13-dimensional binary images, whose pseudo ground truth is pre-computed by ~\cite{yu2018bisenet}. The segmentation loss is given by (\ref{seg_mtrn_eqn}) as follows:
\begin{align}\label{seg_mtrn_eqn}
    L_{Seg{J_i}}(\theta_F,\theta_S) =& CE(x,Y) \nonumber\\
    =& -\sum_{(x,Y)\sim{J_i}}\sum_C y\log{S(F(x_{j,k}))} 
\end{align}
%\begin{multline}\label{seg_tst_eqn}
%L_{Seg({K})}(\theta_F,\theta_{S}) = \\
%\sum_{(x,Y)\sim{D_{mtst}}}\sum_{C}\sum_{(p,y)\sim(x,Y)}y\log{S(F(p))}\\
%+(1-y)\log(1-S(F(p)))
%\end{multline}
where $\theta_S$ denotes the parameters in the face parsing module, $Y$ denotes the ground truth of face parsing masks corresponding to the input, $C$ is the channel of face parsing masks, and $x_{j,k}$ denotes a pixel in our predicted output. $b$ is either a meta-train or meta-test domain depends on the meta-train or meta-test stage.
\subsubsection{Depth Loss}\label{s:deploss}
Similar to RFMetaFAS~\cite{Shao_2020_AAAI}, in order to exploit depth image as domain knowledge to regularize the feature extractor, we apply the L2 loss between the prediction and the pseudo depth ground truth obtained from ~\cite{feng2018joint}. It is given by (\ref{dep_mtrn_eqn}).
\begin{align}\label{dep_mtrn_eqn}
    L_{Dep_b}(\theta_F,\theta_{D}) = \sum_{(x,I)\sim{b}}\|{D(F(x))-I}\|^{2}
\end{align}
%and
%\begin{equation}\label{dep_tst_eqn}
%L_{Dep(D_{mtst})}(\theta_F,\theta_{D}) =
%\sum_{(x,I)\sim{D_{mtst}}}\|{D(F(x))-I}\|^{2}
%\end{equation}
where $\theta_D$ consists of the parameters in the depth estimator, and $I$ denotes the pre-computed depth image.
\subsubsection{Overall Loss}\label{s:allloss}
Finally, the overall objective function is given as follows:
\begin{align}\label{overall_eqn}
L&_{all} = \lambda_{mtrn}L_{J}+\lambda_{mtst}L_{{K}} \nonumber\\
        =& \lambda_{mtrn}(\lambda_{Cls}L_{Cls(J)}+\lambda_{Dep}L_{Dep(J)}+\lambda_{Seg}L_{Seg(J)}+\lambda_{Trip}L_{Trip(J)}) \nonumber\\
        +&\lambda_{mtst}(\lambda_{Cls}L_{Cls({K})}+\lambda_{Dep}L_{Dep({K})}+\lambda_{Seg}L_{Seg({K})}+\lambda_{Trip}L_{Trip({K})})
\end{align}
where $\lambda_{mtrn},\lambda_{mtst},\lambda_{Cls},\lambda_{Dep},\lambda_{Seg},\lambda_{Trip}$ are pre-defined hyperparameters. Here we set $\lambda_{mtrn}=1$, $\lambda_{mtst}=1$, $\lambda_{Cls}=1$, $\lambda_{Dep}=10$, $\lambda_{Seg}=1$, $\lambda_{Trip}=0.5$ in every experiment. To balance the contributions of meta-train and meta-test set, we set $\lambda_{mtrn}=\lambda_{mtst}$. Moreover, the values of $\lambda_{Cls}$, $\lambda_{Dep}$, $\lambda_{Seg}$, and $\lambda_{Trip}$ are modified to balance $L_{Cls}$, $L_{Dep}$, $L_{Seg}$, and $L_{Trip}$ because we find that the results are more stable while they are balanced as we test.

\begin{table*}
\caption{Experimental Comparisons of different face anti-spoofing methods on the four domain generalization experiments.}
\label{tab:exp_result}
\begin{center}
\begin{tabular}{|c||c|c||c|c||c|c||c|c|}
\hline
Method & \multicolumn{2}{c||}{\textbf{O\&C\&I to M}} & \multicolumn{2}{c||}{\textbf{O\&M\&I to C}} & \multicolumn{2}{c||}{\textbf{O\&C\&M to I}} & \multicolumn{2}{c|}{\textbf{I\&C\&M to O}}  \\
{ } & HTER(\%) & AUC(\%) & HTER(\%) & AUC(\%) & HTER(\%) & AUC(\%) & HTER(\%) & AUC(\%) \\
\hline\hline
MADDG\cite{Shao_2019_CVPR} & 17.69 & 88.06 & 24.50 & 84.51 & 22.19 & 84.99 & 27.98 & 80.02 \\
  RFMetaFAS\cite{Shao_2020_AAAI} &  13.89 & 93.98 & 20.27 & 88.16 & 17.30 & 90.48 & 16.45 & 91.16\\
  CCDD\cite{9151064} & 15.42 & 91.13 & 17.41 & 90.12 & 15.87 & 91.47 & 14.72 & 93.08 \\
  PAD-GAN\cite{PAD-GAN} & 17.02 & 90.10 & 19.68 & 87.43 & 20.87 & 86.72 & 25.02 & 81.47 \\
  NAS-FAS\cite{9252183} & 16.85 & 90.42 & 15.21 & 92.64 & 11.63 & \textbf{96.98} & 13.16 & 94.18 \\
  SSDG-M\cite{Jia_2020_CVPR_SSDG} & 16.67 & 90.47 & 23.11 & 85.45 & 18.21 & 94.61 & 25.17 & 81.83 \\
  MT-FAS\cite{MT-FAS-PAMI2021} & 11.67 & 93.09 & 18.44 & 89.67 & 11.93 & 94.95 & 16.23 & 91.18 \\
  D$^2$AM\cite{D2AM-AAAI2021} & 12.70 & 95.66 & 20.98 & 85.58 & 15.43 & 91.22 & 15.27 & 90.87 \\
  ANRL\cite{ANRL-MM2021} & 10.83 & 96.75 & 17.85 & 89.26 & 16.03 & 91.04 & 15.76 & 91.90 \\
  FGHV\cite{FGHV-AAAI2022} & 9.17 & \textbf{96.92} & 12.47 & 93.47 & 16.29 & 90.11 & 13.58 & 93.55 \\
  SSAN-M\cite{SSAN-M-CVPR2022} & 10.42 & 94.76 & 16.47 & 90.81 & 14.00 & 94.58 & 19.51 & 88.17 \\
  \textbf{Ours} & \textbf{7.38} & 96.66 & \textbf{13.2} & \textbf{94.27} & \textbf{8.07} & 96.85 & \textbf{8.75} & \textbf{95.95} \\
\hline
\end{tabular}
\end{center}
\end{table*}

\section{Experimental Results}
\subsection{Experimental Setting}\label{s:settings}
\subsubsection{Datasets}\label{s:datasets}
Here we conduct domain generalization experiments with four public datasets: Oulu-NPU~\cite{oulu-npu}(denoted as O), Idiap Replay-Attack~\cite{6313548}(denoted as I), CASIA-FASD~\cite{6199754}(denoted as C), and MSU-MFSD~\cite{distortion}(denoted as M). These four datasets contain large domain shifts which include variations in illumination, background, imaging devices, and attack types.

The experiment setting follows the work in ~\cite{Shao_2020_AAAI}. We take one of the four source datasets as the unseen testing domain and the other three datasets as training domains in each experiment. Therefore, we conduct four experiments: O\&C\&I to M, O\&M\&I to C, O\&C\&M to I, I\&C\&M to O.

\subsubsection{Implementation Details}\label{s:implementation details}
Our network is implemented with PyTorch. Adam is adopted as the optimizer with momentum of 0.9, weight decay is set to 5e-5, learning rate is 1e-3, and batch size is set to 20 for each domain. For the hyperparameters in the  objective functions, we set $\lambda_1=1$, $\lambda_2=1$, $\lambda_3=1$, $\lambda_4=10$, $\lambda_5=1$, $\lambda_6=0.5$. We concatenate RGB and HSV images as input to our network, which is resized to $256\times256\times6$. The detailed network structure is given in the supplemental material. 

\subsubsection{Evaluation Metrics}\label{s:metrics}
Here we adopt Half Total Error Rate (HTER) and Area Under Curve (AUC) as evaluation metrics. For visualization, we exploit t-SNE~\cite{JMLR:v9:vandermaaten08a} and Grad-CAM~\cite{grad-cam} to show the effect of our approach.

\subsection{Experimental Comparisons}\label{s:comparisons}
In the following section, we compare performance of our approach with several state-of-the-art face anti-spoofing methods, including MADDG~\cite{Shao_2019_CVPR}, and RFMetaFAS~\cite{Shao_2020_AAAI}, CCDD~\cite{9151064}, PAD-GAN~\cite{PAD-GAN}, NAS-FAS~\cite{9252183}, SSDG-M~\cite{Jia_2020_CVPR_SSDG},
MT-FAS~\cite{MT-FAS-PAMI2021}, D2AM~\cite{D2AM-AAAI2021}, ANRL~\cite{ANRL-MM2021}, FGHV~\cite{FGHV-AAAI2022}, and SSAN-M~\cite{SSAN-M-CVPR2022}. Table~\ref{tab:exp_result} demonstrates significant improvements by using the proposed method compared with the baseline RFMetaFAS~\cite{Shao_2020_AAAI} and other state-of-the-art methods.

Note that we use the architecture of MADDG~\cite{Shao_2019_CVPR} as our backbone, which is the same as SSDG-M~\cite{Jia_2020_CVPR_SSDG}. However, SSDG-R~\cite{Jia_2020_CVPR_SSDG} and PatchNet~\cite{Wang_2022_CVPR} adopts ResNet-18~\cite{resnet18} as the backbone, and LMFD-PAD~\cite{Fang_2022_WACV} adopts ResNet-50 as backbone. Thus, they are not included in the experimental comparison because its model size is much larger than MADDG.

\begin{table*}
\caption{Ablation study of using different components in our method. The upper part is about using face parsing information, and the lower part is about one-side triplet loss. For face parsing information, we compare the results of our method with and without using the attention-based skip connection (ASC) and U-net based face parsing module (parsing). For one-side triplet loss, we compare the results of our method with and without one-side triplet loss, without meta learning, and with normal triplet loss.}
\label{tab:ablation}
\begin{center}
\begin{tabular}{|c||c|c||c|c||c|c||c|c|}
\hline
Method & \multicolumn{2}{c||}{\textbf{O\&C\&I to M}} & \multicolumn{2}{c||}{\textbf{O\&M\&I to C}} & \multicolumn{2}{c||}{\textbf{O\&C\&M to I}} & \multicolumn{2}{c|}{\textbf{I\&C\&M to O}} \\
{ } & HTER(\%) & AUC(\%) & HTER(\%) & AUC(\%) & HTER(\%) & AUC(\%) & HTER(\%) & AUC(\%) \\
\hline\hline
Ours w/o parsing    & 10.17 & 94.05 & 20.43 & 88.64 & 13.12 & 94.49 & 17.26 & 90.95 \\
Ours w/o ASC    & 9.31 & 94.30 & 26.68 & 83.72 & 9.13 & 96.46 & 15.01 & 93.06  \\
\hline
Ours w/o one-side trip.    & 9.25 & 95.72 & 25.03 & 83.97 & 12.30 & 95.08 & 13.79 & 93.51 \\
Ours w/o meta    & 10.94 & 95.30 & 31.05 & 75.78 & 27.79 & 79.26 & 29.24 & 77.70 \\
Ours w/ normal trip. & 9.94 & 95.43 & 22.54 & 87.17 & 10.68 & 95.59 & 14.42 & 92.78 \\ 
\hline
\textbf{Ours}    & \textbf{7.38} & \textbf{96.66} & \textbf{13.20} & \textbf{94.27} & \textbf{8.07} & \textbf{96.85} & \textbf{8.75} & \textbf{95.95} \\
\hline
\end{tabular}
\end{center}
\end{table*}

\subsection{Face Parsing Results}\label{s:segmap_exp}
\begin{figure}[t]
\begin{center}
   \includegraphics[width=\linewidth]{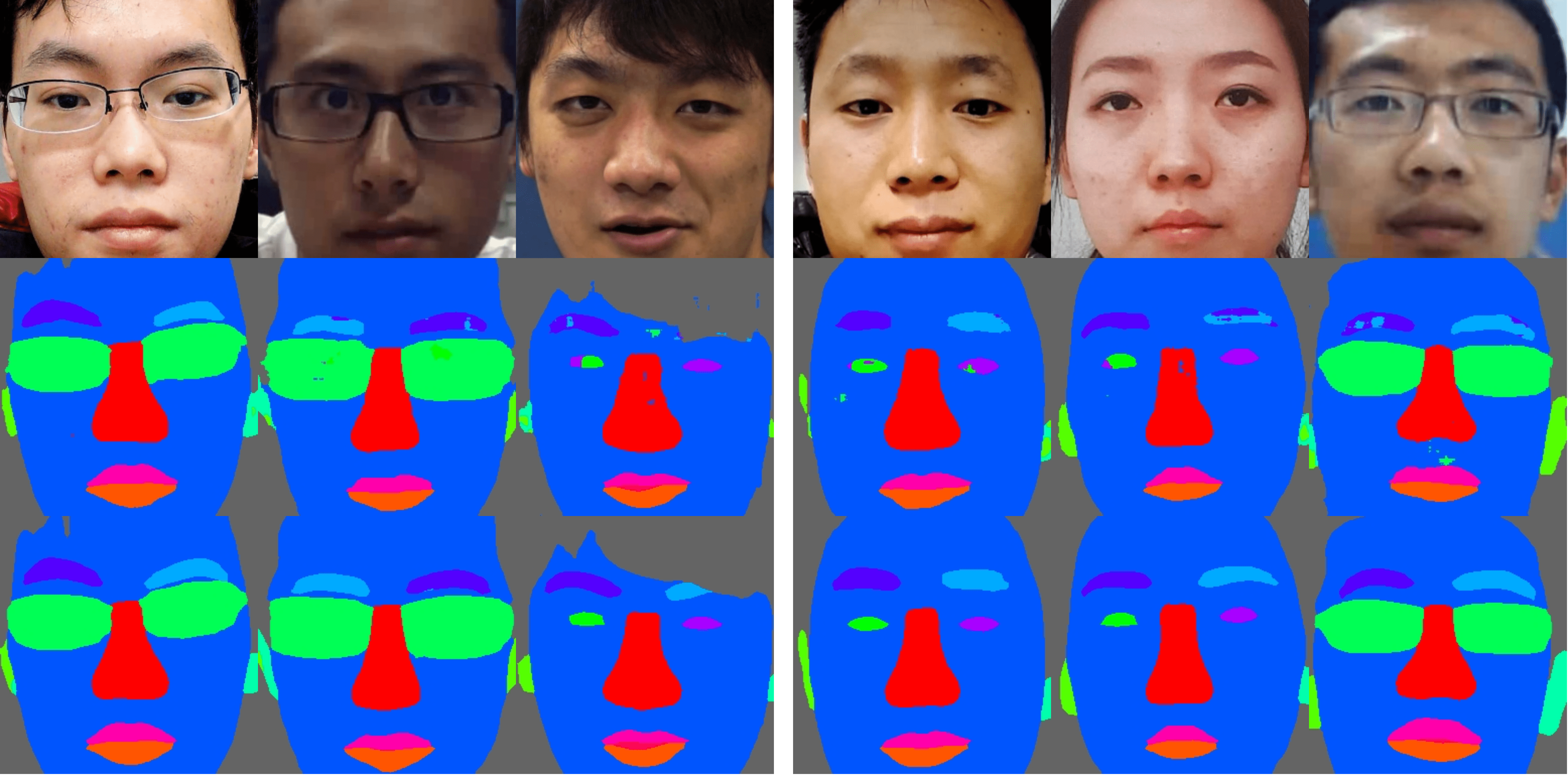}
\end{center}
   \caption{Segmentation output from the face parsing module. From top to bottom: input image, parsing output, and ground truth. The left three columns are live samples, and the right three columns are spoof ones.}
\label{fig:exp-segout}
\end{figure}

We depict visualization of the output of U-net based face parsing module in Fig. ~\ref{fig:exp-segout}. 13 labels are chosen for the face parsing: skin, left/right brow, left/right eye, eyeglasses, left/right ear, nose, mouth, upper/lower lip, and background. The pixel-wise output of semantic facial priors is well performed both for real and spoof images.  

\subsection{Ablation Study}\label{s:ablation study}

\subsubsection{U-net Based Face Parsing Module}
To understand the effect of U-net based face module, here we compare our approach with two different experiment settings: our approach without the U-net face parsing module and the attention-based skip connection. For the frontal one, we want to realize the effect of face parsing for the face anti-spoofing task. For the latter one, we want to discuss whether the face parsing information should be encoded and added back to the main network by the attention-based connection. Table ~\ref{tab:ablation} shows the experimental results of the three settings. It shows that the U-net based face parsing module is effective for regularizing the feature, and the attention-based skip connection(ASC) indeed helps to generalize better in the unseen domains for face anti-spoofing.

\subsubsection{One-Side Triplet Loss with Meta-learning}
In the following section, we perform an experiment to show the effect of the combination of one-side triplet loss and meta-learning. Table ~\ref{tab:ablation} shows results of our approach with and without one-side triplet loss, without meta-learning, and with normal triplet loss instead of one-side triplet loss. It is worth noting that HTER and AUC are improved by over 10\% in O\&M\&I to C experiment. Without one-side triplet loss or meta-learning, the AUCs of O\&M\&I to C degrade dramatically. In our opinion, O\&M\&I to C is the harder experiment because there is no cut attack in source domains, but target domain exists cut attack. Thus, the performance improves with one-side triplet loss or meta-learning means that out model performs much generalized. For O\&C\&I to M, the performance does not decline obviously due to the data of M is comparatively easy. In summary, removing meta-learning from our method and changing one-side triplet loss into ordinary triplet loss cause large degradation of the generalization performance in most protocols.

\begin{figure*}[htpb]
\begin{center}
   \includegraphics[width=0.8\textwidth]{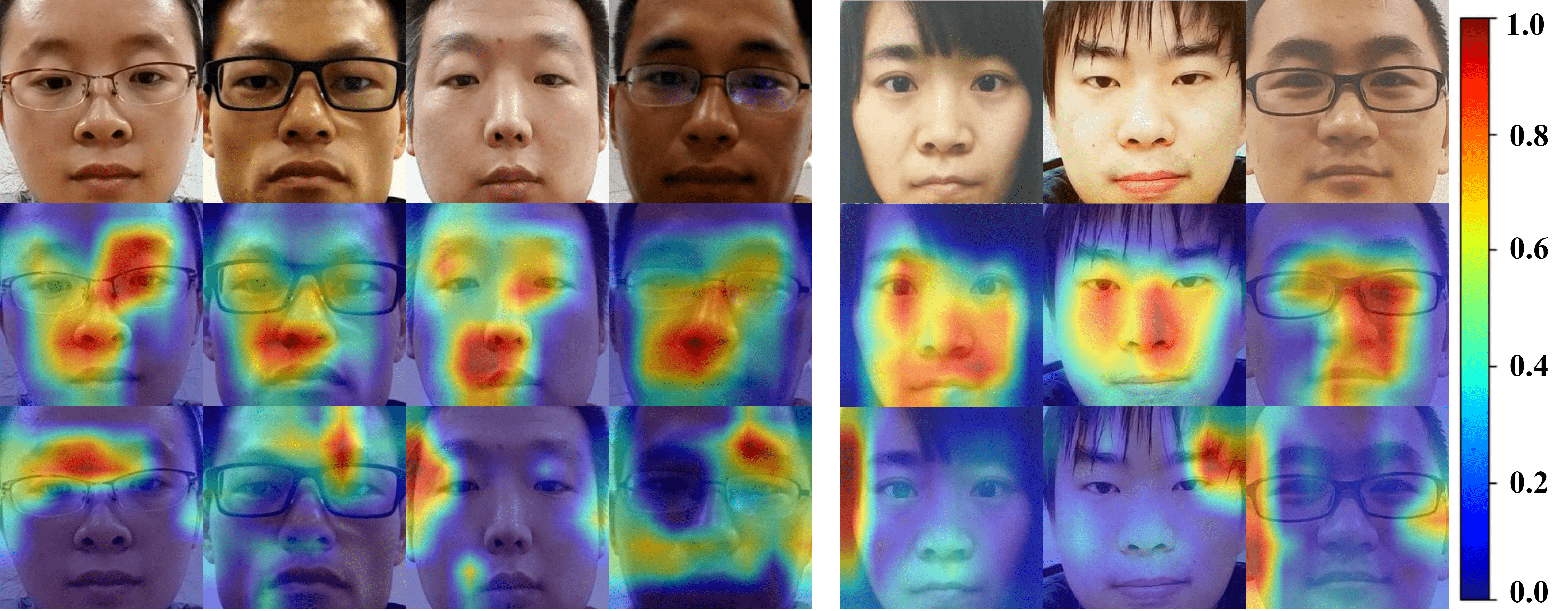}
\end{center}
   \caption{Visualization of Grad-CAM. From top to bottom: input image, our approach, and ours without U-net based face parsing module. The left four columns are live images, and the remaining columns are spoof ones.}
\label{fig:ablation-seg}
\end{figure*}

\begin{figure*}[h]
\begin{center}
%\fbox{\rule{0pt}{2in} \rule{.9\linewidth}{0pt}}
   \includegraphics[width=0.75\textwidth]{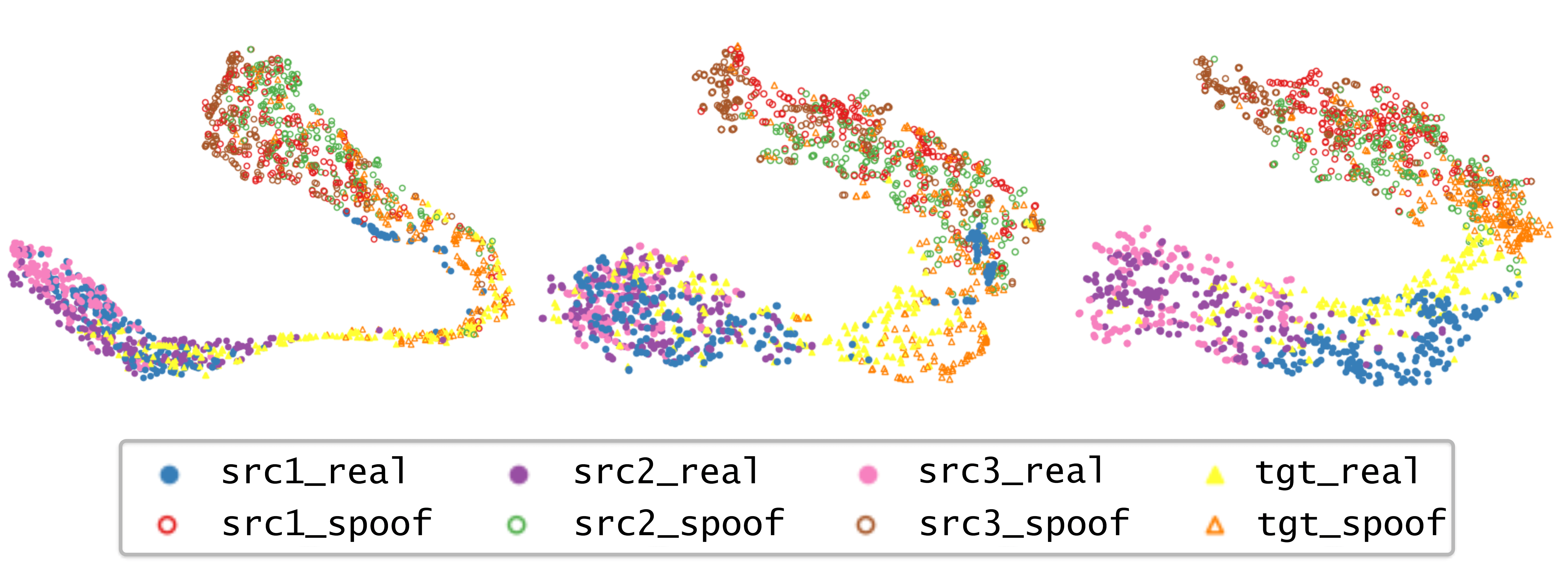}
\end{center}
   \caption{Visualization results of t-SNE based on our approach. From left to right: our approach without triplet loss, ours with normal triplet loss, and our approach. Src means the source domains, and tgt means the target domain.}
\label{fig:tsne-visualize}
\end{figure*}

\subsection{Visualization}\label{s:visualization}
\subsubsection{Grad-CAM Visualization}
In addition to the above experimental results, we depict visualization of some examples with Grad-CAM\cite{grad-cam}, which can localize the important regions that the network concentrated on according to a specific class.

Fig. ~\ref{fig:ablation-seg} shows the results of the two settings: our approach with and without U-net based face parsing module. The three rows are input images, ours, and ours without U-net based face parsing module. The first four columns are live samples, and the others are spoofed ones. We encourage our model to learn discriminative cues from the facial region instead of background because the background contain large variations for different datasets. As Fig. ~\ref{fig:ablation-seg} shows, our network tends to focus on facial parts for both live and spoof images.

\subsubsection{t-SNE Visualization}
t-SNE is adopted for visualizing the effect of one-side triplet loss. Fig. ~\ref{fig:tsne-visualize} shows three settings from left to right: our approach without triplet loss, ours with normal triplet loss, and ours with one-side triplet loss. Here we apply the margin of normal triplet loss as 0.1, which is the same as our beginning setting. As shown in Fig. ~\ref{fig:tsne-visualize}, target domain samples are well discriminated against those with one-side triplet loss. To achieve generalization, we do not separate different domains and thus reduce the domain gaps.

\begin{figure}[h]
\begin{center}
%\fbox{\rule{0pt}{2in} \rule{.9\linewidth}{0pt}}
    \includegraphics[width=0.9\linewidth]{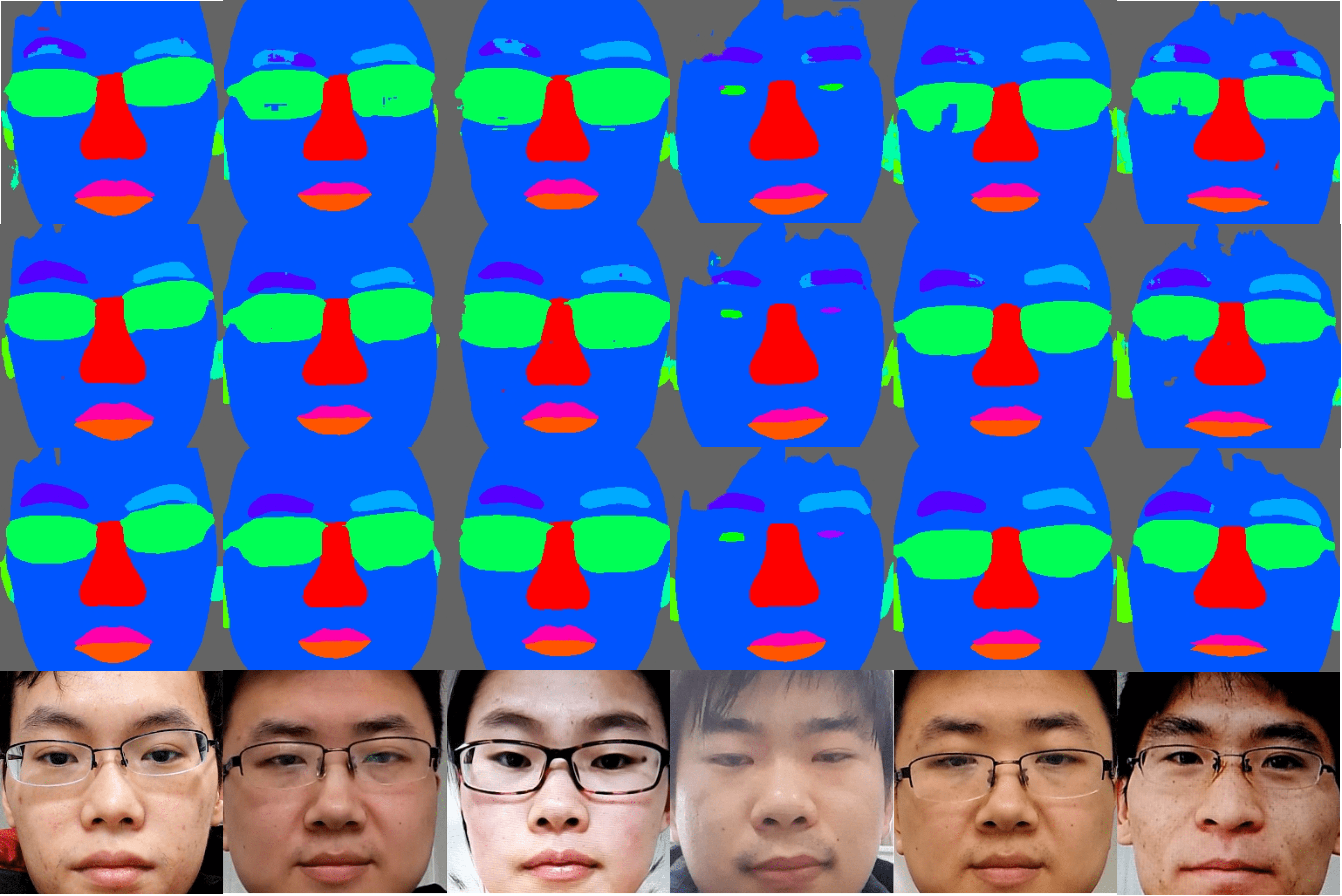}
\end{center}
   \caption{Segmentation output from the face parsing module. Four rows are our approach without attention-based skip connection, our approach, ground truth, and the input image.}
\label{fig:seg-no-branch}
\end{figure}

\subsubsection{Effect of Attention-Based Skip Connection for Face Parsing}\label{supple-seg-no-branch}
We visualize the segmentation output from the face parsing module. Here we mainly focus on comparing our approach to ours without attention-based skip connection. As Fig. \ref{fig:seg-no-branch} shown, adding attention-based skip connection to our approach makes the face parsing results more stable and refined.

\section{Conclusion}\label{s:conclusion}
In this paper, we propose a multi-task framework based on meta learning to improve the model generalization capability for face anti-spoofing. For the face parsing task, a U-net based face parsing module is proposed to learn important face parsing information. For the spoof classification task, a one-side triplet loss is employed to combine the advantages of meta learning and triplet loss. Our experiments on cross-domain generalization for face anti-spoofing demonstrate that our method provides superior performance compared to the state-of-the-art methods on public datasets.

% end
%-----------------------------------------------------------------------%

%\begin{table}
%\caption{An Example of a Table}
%\label{table_example}
%\begin{center}
%\begin{tabular}{|c||c|}
%\hline
%One & Two\\
%\hline
%Three & Four\\
%\hline
%\end{tabular}
%\end{center}
%\end{table}

%\pagebreak

{\small
\bibliographystyle{ieee}
\bibliography{FG2023-anti-spoofing.bib}
}

\end{document}